\documentclass{article}
\usepackage{float}
\title{\Huge{Numeration-automatic sequences}}
\author{Jeroen F. J. Laros}
\date{\today}

\usepackage{./graphs}

\usepackage{amsfonts, amssymb, amsthm}

\newcommand{\monoit}[1]{\texttt{\textit{#1}}}
\newcommand{\spaces}{\begin{picture}(9,0)(0,0)\end{picture}}

\newtheorem{theorem}{Theorem}[subsection]
\newtheorem{lemma}[theorem]{Lemma}
\newtheorem{corollary}[theorem]{Corollary}

\theoremstyle{definition}
\newtheorem{example}[theorem]{Example}
\newtheorem{definition}[theorem]{Definition}

\begin{document}
\maketitle
\newpage

\tableofcontents
\newpage

\section{Introduction}
The Fibonacci substitution \cite{Fogg}, page 51 caught our interest because it 
defines a numeration system and we wondered if there are other substitutions 
having the same property. We present a class of substitutions which generate 
numeration systems. For more information about Automata and formal language
theory, see \cite{Wood} and for more on numeration systems see \cite{Loth}.

\section{Definitions}
First we shall give a couple of definitions which we will use in this document.

\begin{definition}[Finite automaton] \label{def:automaton}
A \emph{finite automaton} 
$A = \{\mathcal{S}, \Delta, \delta, I, F, Y, \varphi\}$ is a tuple in which:
\begin{itemize}
\item $\mathcal{S}$ is the finite set of \emph{states}.
\item $\Delta$ is the finite set of \emph{labels}.
\item $\delta \subseteq \mathcal{S} \times \Delta \times \mathcal{S}$ is the
      collection of \emph{transitions}.
\item $I \subseteq \mathcal{S}$ is the collection of \emph{initial states}.
\item $F \subseteq \mathcal{S}$ is the collection of \emph{final states}.
\item $Y$ the output alphabet.
\item $\varphi$ is a function from $\mathcal{S}$ to $Y$ named the output
      function or exit map.
\end{itemize}
\end{definition}

We represent an automaton by a directed graph with a set of vertices
$\mathcal{S}$ called \emph{states}, a set of edges $\delta$ called
\emph{transitions} and specially marked subsets of states $I$ and $F$, the
\emph{initial} and \emph{final} states.

All through this document we shall take $I = \{\iota\}$ as the only initial
state and $F = \mathcal{S}$ as the collection of final states and we shall
take $\Delta \subset \mathbb{N}$ unless stated otherwise.

Furthermore we shall usually take $Y = \mathcal{S}$ and $\varphi = Id$ as the
output alphabet and the exit map.

Note that the output function applies to the states and not to the labels.

\begin{definition}[Regular language] \label{def:regular_language}
The language $\mathcal{L}(A)$ of a finite automaton is called a 
\emph{regular language}. The language of an automaton is the collection of
strings that are \emph{accepted} by the automaton (all paths in the automaton
that lead from the initial state to a final state).
\end{definition}

\begin{definition}[Regular expression] \label{def:regular_expression}
Let $\mathcal{A}$ be an alphabet. Then a \emph{regular expression} $E$ over 
$\mathcal{A}$ is defined recursively as one of the following types:
\begin{itemize}
\item $\varnothing$.
\item $\epsilon$.
\item $a$, where $a \in \mathcal{A}$.
\item $(E_1 \cup E_2)$, where $E_1$ and $E_2$ are regular expressions and the
      $\cup$ operator denotes a union.
\item $(E_1 \cdot E_2)$, where $E_1$ and $E_2$ are regular expressions and the
      $\cdot$ operator denotes concatenation.
\end{itemize}
\end{definition}

Apart from the types in \ref{def:regular_expression} we shall use some other 
notations.
\begin{itemize}
\item $E^*$, where $E$ is a regular expression and the $^*$ operator denotes the
      union of all powers of $E$, so ($E^* = \cup_{n \in \mathbb{N}}E^n$).
\item $E^+$ is an abbreviation for $E \cdot E^*$.
\end{itemize}

We usually omit the $\cdot$ in a regular expression. The $\cup$ is sometimes
written as `+' or `,'.

Note that a regular expression must be of finite length. Otherwise we call it
an infinite automaton.

\begin{definition}[Substitution] \label{def:substitution}
A \emph{substitution} $\sigma$ is a function from an alphabet $\mathcal{A}$ to 
$\mathcal{A}^* - \{\epsilon\}$ of nonempty finite words on $\mathcal{A}$. It
extends to a substitution on $\mathcal{A}^*$ by concatenation. So 
$\sigma(w w') = \sigma(w) \sigma(w')$. We set $\sigma(\epsilon) = \epsilon$, 
with $\epsilon$ being the empty word.
\end{definition}

\begin{definition}[$n$-word] \label{def:n-word}
An \emph{$n$-word} of a substitution $\sigma$ is the unique word 
$\sigma^n(\iota)$, $\iota \in \mathcal{A}$ being the initial letter.
\end{definition}

Sometimes we need to refer to an element in an $n$-word. We use the notation 
$\sigma^n(\iota)_i$ when we want to refer to the $i$-th element of 
$\sigma^n(\iota)$.

\begin{definition}[Fixed point] \label{def:fixed_point}
A \emph{fixed point} of a substitution $\sigma$ is an infinite sequence $u$ 
with $\sigma(u) = u$.
\end{definition}

\begin{definition}[$k^\mathrm{max}$] \label{def:k-max}
Let $\Delta \subset \mathbb{N}$ be the collection of labels. Define 
$k^\mathrm{max} = |\Delta| - 1$.

Equivalently, let $\sigma$ be a substitution. Define $k^\mathrm{max}$ as
the greatest length of the images in $\sigma$ subtracted by 1.
\end{definition}

\begin{definition}[Numeration system] \label{def:numeration_system}
In general a \emph{numeration system} is a strictly increasing sequence 
$U = (U_i)_{i \in \mathbb{N}}$ such that
\begin{itemize}
\item $U_0 = 1$ (to represent all $n \in \mathbb{N}$),
\item $\mathrm{sup}\frac{U_{i + 1}}{U_i} < \infty$ (to have a finite alphabet 
      of digits).
\end{itemize}
\end{definition}

An expansion of an integer $n \in \mathbb{N}$ in such a numeration system is
a finite sequence $(a_i)_{k \ge i \ge 0}$ such that 
$n = \sum_{i = k}^0 a_i U_i$. We write this expansion as $a_k \ldots a_0$, the
most significant bit is in the first position. Note that there are more than
one expansions in general, but one of them is called the \emph{normal} or
\emph{greedy} representation. We shall discuss this in Section \ref{sec:ea}.

It is quite natural to express the expansion of $U_i$ as
$1\underbrace{00\ldots0}_i$. It is also common practice to give the following
restriction $0 \le a_i \le \lceil\mathrm{sup}\frac{U_{n + 1}}{U_n}\rceil$.

In general more than one expansion can be found for an integer.

\begin{definition}[Full numeration system] \label{def:full_numeration_system}
A \emph{full numeration system} is a numeration system that has the extra
property: 
\begin{itemize}
\item If $A = a_k a_{k - 1} \ldots a_0$ and $B = b_k b_{k - 1} \ldots b_0$, 
with $a_i + b_i \le U_i$ for all $i$, then the sum $A + B$ equals 
$(a_k + b_k) (a_{k - 1} + b_{k - 1}) \ldots (a_0 + b_0)$.
\end{itemize}
\end{definition}

For more on numeration systems, see \cite{Loth}, chapter 7 and in particular
Section 7.3.

\section{Equivalence between substitutions and automata}
\subsection{Substitutions in general}
A substitution $\sigma$ on an alphabet $\mathcal{A}$ defines an automaton in 
the following way:
\begin{itemize}
\item Let $\mathcal{S} = \mathcal{A}$ be the collection of states.
\item Add a transition from state $a$ to state $b$ ($a,b \in \mathcal{S}$) 
labeled $i$ if $b$ occurs in $\sigma(a)$ at position ($i$ + 1).
\item Let $\iota$ be the initial state.
\item Let all states in $\mathcal{S}$ be final states.
\end{itemize}

Every automaton in turn defines a regular language (by definition).
\begin{lemma}[Automata and substitutions] \label{lem:automata_substitutions}
Let $\sigma$ be a substitution which is in bijection with an automaton $A$,
let $\mathcal{L}(A)$ be the language of the automaton and let $i > 0$. Then 
$\sigma^n(\iota)_i$ is the state the automaton will be in after it is fed with 
the $i$-th word of length $n$ of its (lexicographical ordened) input language.
\end{lemma}

\begin{proof}
If we write the automaton as a computation tree, the equivalence is easier to 
see. We find $\sigma^n(\iota)$ by reading all states at depth $n$ in the tree. 
Let us assume that all words in $\mathcal{L}(A)$ of length $n$ are in order.
Now we look at words of length $n + 1$. Every word of length $n + 1$ comes from
a word of length $n$. At the $i$-th word of length $n$, the automaton is in 
state $\sigma^n(\iota)_i$. The words of length $n + 1$ that are possible by 
extending the word of length $n$ are exactly those which are extended with the 
elements from $\sigma^n(\iota)_i$. These are added in order.
\end{proof}

\begin{example} \label{ex:automata_substitutions}
In Figure \ref{fig:automata_substitutions} we see an automaton $A$ in bijection
with substitution $\sigma$.

\begin{figure}[H]
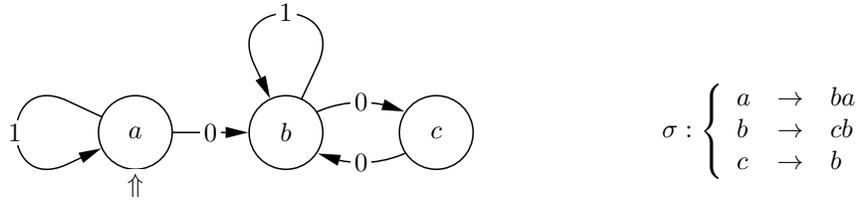

\vbox{
\ \\
\ \\
\begin{eqnarray*}
\spaces
\sigma: \left\{ \begin{array}{lll}
a &\rightarrow& ba\\
b &\rightarrow& cb\\
c &\rightarrow& b
\end{array} \right.
\end{eqnarray*}
\begin{graph}(0, 0)(-4, -1)
  \graphnodecolour{1}
  \graphnodesize{1}
  \roundnode{s1}(-2, 0) \nodetext{s1}(0, 0){$a$}
  \roundnode{s2}(0, 0)  \nodetext{s2}(0, 0){$b$}
  \roundnode{s3}(2, 0)  \nodetext{s3}(0, 0){$c$}

  \diredge{s1}{s2} \edgetext{s1}{s2}{0}
  \dirloopedge{s1}{50}(-1, 0) \freetext(-3.6, 0){1}
  \dirbow{s2}{s3}{.2} \bowtext{s2}{s3}{.2}{0}
  \dirloopedge{s2}{50}(0, 1) \freetext(0, 1.6){1}
  \dirbow{s3}{s2}{.2} \bowtext{s3}{s2}{.2}{0}

  \freetext(-2, -0.7){$\Uparrow$}
\end{graph}}
\caption{An automaton in bijection with substitution $\sigma$}
\label{fig:automata_substitutions}
\end{figure}
The computation tree of $A$ accepts the same language as the automaton.

\begin{figure}[H]
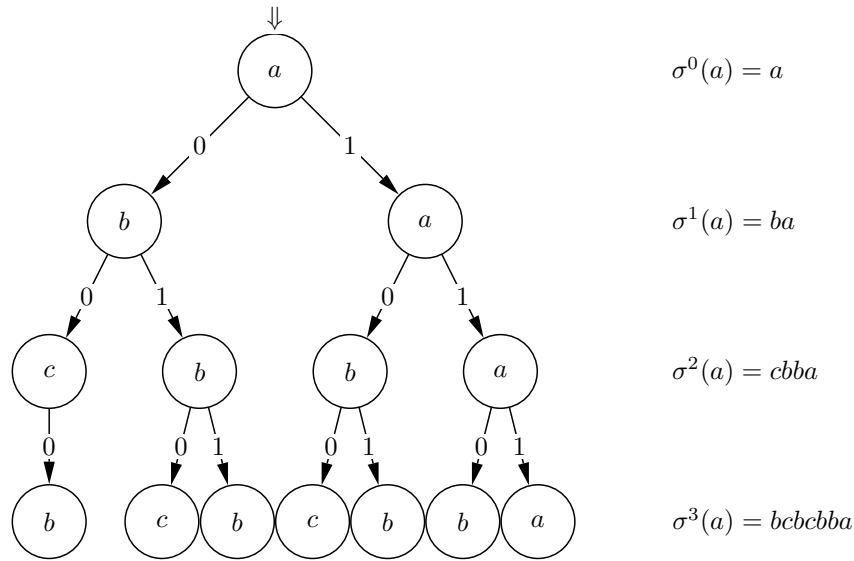

\begin{graph}(0, 7.5)(-4, -2.5)
  \graphnodecolour{1}
  \graphnodesize{1}
  \roundnode{s1}(0, 4) \nodetext{s1}(0, 0){$a$} 
  \freetext(6.02, 4){$\sigma^0(a) = a$}

  \roundnode{s2}(-2, 2)  \nodetext{s2}(0, 0){$b$} 
  \roundnode{s3}(2, 2)  \nodetext{s3}(0, 0){$a$}
  \freetext(6.1, 2){$\sigma^1(a) = ba$}

  \roundnode{s4}(-3, 0)  \nodetext{s4}(0, 0){$c$} 
  \roundnode{s5}(-1, 0)  \nodetext{s5}(0, 0){$b$}
  \roundnode{s6}(1, 0)  \nodetext{s6}(0, 0){$b$}
  \roundnode{s7}(3, 0)  \nodetext{s7}(0, 0){$a$}
  \freetext(6.25, 0){$\sigma^2(a) = cbba$}

  \roundnode{s8}(-3, -2)  \nodetext{s8}(0, 0){$b$} 
  \roundnode{s9}(-1.5, -2)  \nodetext{s9}(0, 0){$c$}
  \roundnode{s10}(-0.5, -2) \nodetext{s10}(0, 0){$b$}

  \roundnode{s11}(0.5, -2)  \nodetext{s11}(0, 0){$c$}
  \roundnode{s12}(1.5, -2)  \nodetext{s12}(0, 0){$b$}
  \roundnode{s13}(2.5, -2)  \nodetext{s13}(0, 0){$b$}
  \roundnode{s14}(3.5, -2)  \nodetext{s14}(0, 0){$a$}
  \freetext(6.48, -2){$\sigma^3(a) = bcbcbba$}

  \diredge{s1}{s2} \edgetext{s1}{s2}{0}
  \diredge{s1}{s3} \edgetext{s1}{s3}{1}

  \diredge{s2}{s4} \edgetext{s2}{s4}{0}
  \diredge{s2}{s5} \edgetext{s2}{s5}{1}
  \diredge{s3}{s6} \edgetext{s3}{s6}{0}
  \diredge{s3}{s7} \edgetext{s3}{s7}{1}

  \diredge{s4}{s8} \edgetext{s4}{s8}{0}
  \diredge{s5}{s9} \edgetext{s5}{s9}{0}
  \diredge{s5}{s10} \edgetext{s5}{s10}{1}
  \diredge{s6}{s11} \edgetext{s6}{s11}{0}
  \diredge{s6}{s12} \edgetext{s6}{s12}{1}
  \diredge{s7}{s13} \edgetext{s7}{s13}{0}
  \diredge{s7}{s14} \edgetext{s7}{s14}{1}

  \freetext(0, 4.7){$\Downarrow$}
\end{graph}
\caption{A computation tree in bijection with $\sigma$}
\label{fig:automata_tree_substitutions}
\end{figure}
The language accepted by both automata is 
$1^* + (1^* 0 (1 + 00)^* (0 + \epsilon))$. The first few elements of this
language are:
\begin{displaymath}
\mathcal{L}(A) = \{\underbrace{\epsilon,}_0
\underbrace{0, 1,}_1
\underbrace{00, 01, 10, 11,}_ 2
\underbrace{000, 010, 011, 100, 101, 110, 111,}_3 \ldots\}
\end{displaymath}

If for example we feed all strings of length 3 of its input language, we shall
end respectively in the states b, c, b, c, b, b, a, which is exactly the string
$\sigma^3(a)$.
\end{example}

\subsection{Substitutions with fixed points} 
Of course Lemma \ref{lem:automata_substitutions} is also valid for 
substitutions with a fixed point. If we have a fixed point, the resulting 
automaton will accept leading zeroes in its input language; leading zeroes do
not change the state of the automaton, so $0^*w$ ends in the same state as $w$.
This is one of the conditions to make a numeration system. From now on we shall
only look at substitutions with a fixed point. We shall also ignore words in 
$\mathcal{L}(A)$ with leading zeroes.

\begin{corollary}[Automata and substitutions with a fixed point] 
\label{cor:automata_substitutions_fixed_point}
Let $\sigma$ be a substitution which is in bijection with an automaton
$A$, let $u$ be the fixed point of the substitution and let $\mathcal{L}(A)$ 
be the language of the automaton. Then $u_i$ is the state the automaton 
will be in after it is fed with the $i$-th word of its input language.
\end{corollary}

\begin{proof}
This follows directly from Lemma \ref{lem:automata_substitutions}.
\end{proof}

\begin{example}
\begin{figure}[H]
\vbox{
\ \\
\ \\
\begin{eqnarray*}
\spaces
\sigma: \left\{ \begin{array}{lll}
a &\rightarrow& ab\\
b &\rightarrow& cb\\
c &\rightarrow& b
\end{array} \right.
\end{eqnarray*}
\begin{graph}(0, 0)(-4, -1)
  \graphnodecolour{1}
  \graphnodesize{1}
  \roundnode{s1}(-2, 0) \nodetext{s1}(0, 0){$a$}
  \roundnode{s2}(0, 0)  \nodetext{s2}(0, 0){$b$}
  \roundnode{s3}(2, 0)  \nodetext{s3}(0, 0){$c$}

  \dirloopedge{s1}{50}(-1, 0) \freetext(-3.6, 0){0}
  \diredge{s1}{s2} \edgetext{s1}{s2}{1}
  \dirbow{s2}{s3}{.2} \bowtext{s2}{s3}{.2}{0}
  \dirloopedge{s2}{50}(0, 1) \freetext(0, 1.6){1}
  \dirbow{s3}{s2}{.2} \bowtext{s3}{s2}{.2}{0}

  \freetext(-2, -0.7){$\Uparrow$}
\end{graph}}
\caption{An automaton with a fixed point}
\label{fig:automata_fixpoint_substitutions}
\end{figure}
The fixed point of $\sigma$ shown in Figures 
\ref{fig:automata_fixpoint_substitutions} and 
\ref{fig:automata_tree_fixpoint_substitutions} with $a$ as the initial letter 
is:\\
\\
\monoit{u = abcbbcbcbbcbbcbcbbcbcbbcbbcbcbbcbbcbcbbcbcbbcbbcbcbbc\ldots}\\
\\
and\\
\\
$\mathcal{L}(A) = \{\epsilon, 0, 1, 10, 11, 100, 110, 111, 1000, 1001, 1100, 
1110, 1111, 10000, \ldots\}$

\begin{figure}[H]
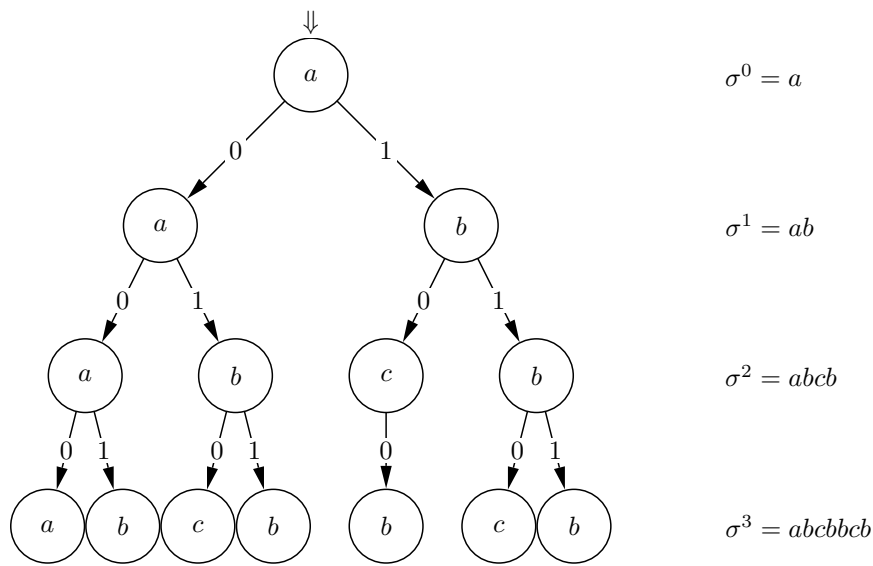

\begin{graph}(0, 7.5)(-4, -2.5)
  \graphnodecolour{1}
  \graphnodesize{1}
  \roundnode{s1}(0, 4) \nodetext{s1}(0, 0){$a$} 
  \freetext(6.02, 4){$\sigma^0 = a$}

  \roundnode{s2}(-2, 2)  \nodetext{s2}(0, 0){$a$} 
  \roundnode{s3}(2, 2)  \nodetext{s3}(0, 0){$b$}
  \freetext(6.1, 2){$\sigma^1 = ab$}

  \roundnode{s4}(-3, 0)  \nodetext{s4}(0, 0){$a$} 
  \roundnode{s5}(-1, 0)  \nodetext{s5}(0, 0){$b$}
  \roundnode{s6}(1, 0)  \nodetext{s6}(0, 0){$c$}
  \roundnode{s7}(3, 0)  \nodetext{s7}(0, 0){$b$}
  \freetext(6.25, 0){$\sigma^2 = abcb$}

  \roundnode{s8}(-3.5, -2)  \nodetext{s8}(0, 0){$a$} 
  \roundnode{s9}(-2.5, -2)  \nodetext{s9}(0, 0){$b$}
  \roundnode{s10}(-1.5, -2) \nodetext{s10}(0, 0){$c$}
  \roundnode{s11}(-0.5, -2)  \nodetext{s11}(0, 0){$b$}
  \roundnode{s12}(1, -2)  \nodetext{s12}(0, 0){$b$}
  \roundnode{s13}(2.5, -2)  \nodetext{s13}(0, 0){$c$}
  \roundnode{s14}(3.5, -2)  \nodetext{s14}(0, 0){$b$}
  \freetext(6.48, -2){$\sigma^3 = abcbbcb$}

  \diredge{s1}{s2} \edgetext{s1}{s2}{0}
  \diredge{s1}{s3} \edgetext{s1}{s3}{1}
  
  \diredge{s2}{s4} \edgetext{s2}{s4}{0}
  \diredge{s2}{s5} \edgetext{s2}{s5}{1}
  \diredge{s3}{s6} \edgetext{s3}{s6}{0}
  \diredge{s3}{s7} \edgetext{s3}{s7}{1}

  \diredge{s4}{s8} \edgetext{s4}{s8}{0}
  \diredge{s4}{s9} \edgetext{s4}{s9}{1}
  \diredge{s5}{s10} \edgetext{s5}{s10}{0}
  \diredge{s5}{s11} \edgetext{s5}{s11}{1}
  \diredge{s6}{s12} \edgetext{s6}{s12}{0}
  \diredge{s7}{s13} \edgetext{s7}{s13}{0}
  \diredge{s7}{s14} \edgetext{s7}{s14}{1}

  \freetext(0, 4.7){$\Downarrow$}
\end{graph}
\caption{A computation tree with a fixed point}
\label{fig:automata_tree_fixpoint_substitutions}
\end{figure}
\end{example}

\section{Numeration-automatism}
Although Corollary \ref{cor:automata_substitutions_fixed_point} is valid for 
all automata that are in bijection with a substitution which have a fixed point
(that means a fairly large group of automata), it is not as useful as it might 
seem at first glance. The problem is that in general we can not give the $n$-th
word of a language $\mathcal{L}$ a priory.

However, there is an obvious class of automata for which we can give the $n$-th
word, the so-called $k$-automata. In this class each letter has a substitution
word of length $k$. For this class of automata the $n$-th word of its input 
language is the $k$-base expansion of $n$.

There are also some non-$k$-automata for which we can describe the $n$-th word 
without much calculation. This class has the property that we can define a 
numeration system in which the expansion of $n$ is the $n$-th word of 
$\mathcal{L}(A)$. We shall refer to these substitutions as numeration-automatic 
substitutions. In general however, a substitution is neither $k$- nor 
numeration-automatic. Sometimes there even exist numbers $n$ for which we can 
not find a valid expansion.

\subsection{The Fibonacci substitution}
The Fibonacci substitution and its automaton can be seen in Figure 
\ref{fig:fibonacci}.

\begin{figure}[H]
\vbox{
\begin{eqnarray*}
\spaces
\sigma: \left\{ \begin{array}{lll}
a &\rightarrow& ab\\
b &\rightarrow& a
\end{array} \right.
\end{eqnarray*}
\begin{graph}(0, 1)(-4, -1.75)
  \graphnodecolour{1}
  \graphnodesize{1}
  \roundnode{s1}(-2, 0) \nodetext{s1}(0, 0){$a$}
  \roundnode{s2}(0, 0)  \nodetext{s2}(0, 0){$b$}

  \dirloopedge{s1}{50}(-1, 0) \freetext(-3.6, 0){0}
  \dirbow{s1}{s2}{.2} \bowtext{s1}{s2}{.2}{1}
  \dirbow{s2}{s1}{.2} \bowtext{s2}{s1}{.2}{0}

  \freetext(-2, -0.7){$\Uparrow$}
\end{graph} $(0 + 10)^* (\epsilon + 1)$}
\caption{The Fibonacci automaton}
\label{fig:fibonacci}
\end{figure}

If we take $a$ as the initial letter the substitution gives the following fixed
point:\\
\\
\monoit{u = abaababaabaababaababaabaababaabaababaababaabaababaaba\ldots}\\
\\
The first elements of the language the automaton defines are:

\begin{displaymath}
\mathcal{L}(A) = \{\epsilon, 0, 1, 10, 100, 101, 1000, 1001, 1010, 10000, 
10001, 10010, 10100, \ldots\}
\end{displaymath}
This language has no consecutive ones. We will call the $(n + 2)$-th word in 
this sequence the Zeckendorf expansion of an integer $n$.

\begin{definition}[The Fibonacci sequence]
Let $(F_i)_{i \in \mathbb{N}}$ be the sequence of integers defined by $F_0 = 1,
F_1 = 2$ and for any integer $i > 1, F_{i + 1} = F_{i - 1} + F_i$.
\end{definition}

\begin{definition}[The Zeckendorf expansion]
If $n = \sum_{i = 0}^k a_i F_i$ with $a_k = 1, a_i \in \{0, 1\}$ and
$\forall (i < k) \{a_i a_{i + 1} = 0\}$, we say that
Zeck$(n) = a_k a_{k - 1} \ldots a_0 \in \{0, 1\}^{k + 1}$ is the Zeckendorf
expansion of the integer $n$.
\end{definition}

The Zeckendorf algorithm is actually an instance of the greedy, or Euclidean
algorithm. We shall discuss this in more detail in Section \ref{sec:ea}.

If we write the partitions of
$\mathbb{N}$ as $\mathbb{F}_a$ and $\mathbb{F}_b$,
\begin{eqnarray*}
  \mathbb{F}_a &=& \{n \in \mathbb{N}, \mathrm{Zeck}(n) \in \{0, 1\}^* 0\}\\
  \mathbb{F}_b &=& \{n \in \mathbb{N}, \mathrm{Zeck}(n) \in \{0, 1\}^* 1\}
\end{eqnarray*}
Then
\begin{eqnarray*}
\;\;\;\;\;\;\;\;\;\;
  \mathbb{F}_a &=& \{0, 2, 3, 5, 7, 8, 10, 11, 13, 15, \ldots\}\\
  \mathbb{F}_b &=& \{1, 4, 6, 9, 12, 14, 17, 19, 22, 25, \ldots\}.
\end{eqnarray*}
Hence we get an $a$ at position $n$ if $n$ ends with a 0 in the
Zeckendorf expansion, we get an $b$ otherwise.\\
\\
So if we want to calculate the $n$-th word of the language this automaton 
defines, we only have to expand $n$ with the Zeckendorf algorithm.

\subsection{Fibonacci's `brother'}
In Figure: \ref{fig:fibonacci_brother} we see another substitution that is 
numeration-automatic.

\begin{example}
\begin{figure}[H]
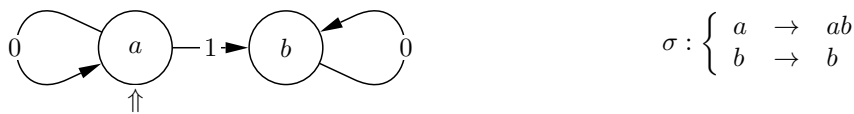

\begin{eqnarray*}
\spaces
\sigma: \left\{ \begin{array}{lll}
a &\rightarrow& ab\\
b &\rightarrow& b
\end{array} \right.
\end{eqnarray*}
\begin{graph}(0, 1)(-4, -1.75)
  \graphnodecolour{1}
  \graphnodesize{1}
  \roundnode{s1}(-2, 0) \nodetext{s1}(0, 0){$a$}
  \roundnode{s2}(0, 0)  \nodetext{s2}(0, 0){$b$}

  \dirloopedge{s1}{50}(-1, 0) \freetext(-3.6, 0){0}
  \diredge{s1}{s2} \edgetext{s1}{s2}{1}
  \dirloopedge{s2}{50}(1, 0) \freetext(1.6, 0){0}

  \freetext(-2, -0.7){$\Uparrow$}
\end{graph} $0^* (\epsilon + 10^*)$
\caption{A Fibonacci-like automaton}
\label{fig:fibonacci_brother}
\end{figure}

The numeration system for this substitution is based on the sequence 
$(n)_{n \in \mathbb{N}^+}$. So the first few expansions are as follows:
\begin{verbatim}
0 -> 0
1 -> 1
2 -> 10
3 -> 100
4 -> 1000
 ...
\end{verbatim}
\end{example}

\section{A generalization}
Can we find a more general class of automata that have similar properties? We 
will show that the answer is yes for a (possibly small) class of substitutions.
This class can be found by calculating the sequence on which the expansion is 
based from the substitution itself. If we look at the Fibonacci substitution, 
we see that for each word $\sigma_n$:
\begin{eqnarray*}
|\sigma_n|_a &=& |\sigma_{n - 1}|_a + |\sigma_{n - 1}|_b\\
|\sigma_n|_b &=& |\sigma_{n - 1}|_a
\end{eqnarray*}
Or in matrix form:
\begin{displaymath} F = \left( \begin{array}{cc}
1 & 1\\
1 & 0
\end{array} \right) \end{displaymath}
If we now define the initial matrix as:
\begin{displaymath} \left( \begin{array}{c}
1 \\
0 
\end{array} \right) \end{displaymath}
and multiply it from the left repeatedly with the (2 $\times$ 2) matrix $F$, we
obtain:
\begin{displaymath} 
\left( \begin{array}{c}
1 \\
0 
\end{array} \right),
\left( \begin{array}{c}
1 \\
1 
\end{array} \right),
\left( \begin{array}{c}
2 \\
1 
\end{array} \right),
\left( \begin{array}{c}
3 \\
2 
\end{array} \right),
\left( \begin{array}{c}
5 \\
3 
\end{array} \right),
\left( \begin{array}{c}
8 \\
5 
\end{array} \right),
\left( \begin{array}{c}
13 \\
8 
\end{array} \right), \ldots
\end{displaymath}
and if we add the elements of the matrices we obtain the Fibonacci sequence 
(note that the elements of the vectors also form the Fibonacci sequence).\\
\\
The matrix for the `brother' of the Fibonacci sequence is
\begin{displaymath} \left( \begin{array}{cc}
1 & 0\\
1 & 1
\end{array} \right), \end{displaymath}
which applied to the initial matrix yields the sequence 
${1 \choose n}_{n \in \mathbb{N}}$.

By using this method we find a numeration system for any automaton. And by 
using a greedy generalized Zeckendorf expansion we can cover all $k$-automata, 
the Fibonacci automaton and its `brother'.

\begin{lemma}[Substitutions and numeration systems] \label{lem:subst-numsys}
If a substitution $\sigma$ has a fixed point, then the sequence
$|\iota|, |\sigma(\iota)|, |\sigma^2(\iota)|, \ldots$ is a numeration system.
\end{lemma}

\begin{proof}
A sequence of integers is a numeration system when it complies to the 
restrictions of Definition \ref{def:numeration_system}. The first demand is 
adhered to because $\iota$ is a letter, so $|\iota| = 1$. The second demand
is adhered to as well, because 
$\mathrm{sup}\frac{U_{i + 1}}{U_i} \le k^{\mathrm{max}}$ and the sequence is 
infinite, otherwise there would be no fixed point.
\end{proof}

Note that this proof also holds for most substitutions, as long as 
$\sigma_n(\iota) < \sigma^{n + 1}(\iota)$.

\subsection{The expansion algorithm} \label{sec:ea}
From a substitution we can extract a numeration system, but we still need an
expansion algorithm to generate $\mathcal{L}(A)$. 

\begin{definition}[General Zeckendorf expansion] \label{def:greedy}
Let $U$ be a full numeration system. If 
$n = \sum_{i = 0}^k a_i U_i$, with 
$0 \le a_i \le k^{\mathrm{max}}$ for $i = 0, 1, \ldots, k - 1$ and 
$a_k > 0$, we say that $a_k a_{k - 1} \ldots a_0$ is the expansion of
$n$ in the $U$ numeration system.
\end{definition}
This algorithm is also known as the greedy or Euclidean algorithm. Hollander 
\cite{Hol} has described the class of recurrent functions which describe a 
numeration system of which the Euclidean expansion is recognized by automata. 
This however is not the entire class of numeration-automata. We shall discuss 
this class in Theorem \ref{thm:fsa}.

\paragraph{Automatic expansion}
Let $U$ be a numeration system, let $A$ be an automaton and let $A_m$ be 
the state in which the automaton will be after reading the first $m$ letters 
of an input word. Let $t(A_m)$ be the set of outgoing transitions of state 
$A_m$. Assume that $t(A_m) = \{0, 1, \ldots, |t(A_m)| - 1\}$. If we can write 
any integer $n \ge 0$ as $n = \sum_{i = 0}^k a_i U_i$ with 
$a_i \in t(A_i)$ for $i = 0, 1, \ldots, k - 1$ and $a_k > 0$
in a unique way such that the automaton accepts the expansion, then 
Auto($n$) = $a_k a_{k - 1} \ldots a_0$ is said to be the automatic expansion 
of the integer $n$.

Since we are only interested in a mapping from $\mathbb{N}^+$ to the regular
language the automaton accepts, it makes sense to look at it this way: does
the automatic expansion of an integer represent the integer itself?

We have devised an algorithm that should answer that question for any
substitution:

\begin{definition}[Automatic expansion] \label{def:autoexpand}
The automatic expansion is an extension of the standard greedy algorithm.
\begin{enumerate}
\item Given $n$ and the numeration system $(U_i)_{i \in \mathbb{N}}$. Let 
      $U_i$ be the largest element in $U$ such that $n \le U_i (|t(A_0)| - 1)$.
      Let $m = 1$.
\item Let $a_i$ be the largest element in $t(A_m)$ such that 
      $a_i U_i \le n$ \label{ae:it:two}
\item Replace $n$ by $n - a_i U_i$, $i$ with $i - 1$, $m$ with $m + 1$ and 
      repeat step \ref{ae:it:two} until $n = 0$.
\end{enumerate}
\end{definition}
If this algorithm fails, the number $n$ could not be expanded.

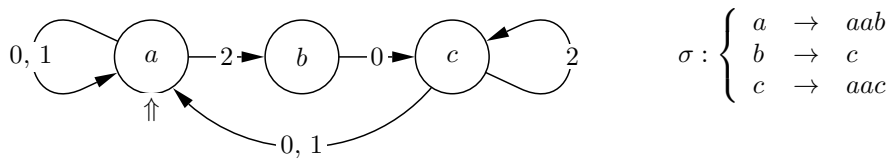
\begin{example} \label{ex:non-semi}
\begin{figure}[H]
\vbox{
\begin{eqnarray*}
\spaces
\sigma: \left\{ \begin{array}{lll}
a &\rightarrow& aab\\
b &\rightarrow& c\\
c &\rightarrow& aac
\end{array} \right.
\end{eqnarray*}
\begin{graph}(0, 1)(-4, -2)
  \graphnodecolour{1}
  \graphnodesize{1}
  \roundnode{s1}(-2, 0) \nodetext{s1}(0, 0){$a$}
  \roundnode{s2}(0, 0)  \nodetext{s2}(0, 0){$b$}
  \roundnode{s3}(2, 0)  \nodetext{s3}(0, 0){$c$}

  \dirloopedge{s1}{50}(-1, 0) \freetext(-3.6, 0){0, 1}
  \diredge{s1}{s2} \edgetext{s1}{s2}{2}
  \diredge{s2}{s3} \edgetext{s2}{s3}{0}
  \dirbow{s3}{s1}{.3} \bowtext{s3}{s1}{.3}{0, 1}
  \dirloopedge{s3}{50}(1, 0) \freetext(3.6, 0){2}

  \freetext(-2, -0.7){$\Uparrow$}
\end{graph}$(0 + 1 + (202^* (0 + 1)))^* 202^* + 2 + \epsilon$}
\caption{A numeration-automaton}
\label{fig:semi1}
\end{figure}
The substitution shown in Figure \ref{fig:semi1} defines the following 
numeration system:
\begin{displaymath}
\{1, 3, 7, 17, 43, 109, 275, 693, 1747, 4405, 11107, \ldots \}
\end{displaymath}
Suppose we want to expand the decimal number 41.

$U_i = 17$, this makes $i = 3$. The expansion goes as follows:

The first element is 2 because $2 \cdot 17 \le 41 < 3 \cdot 17$, this leaves 
$41 - (2 \cdot 17) = 7$ to be expanded. Go to state $b$ (follow label 2).

The second element is 0 because this is the only transition going out of 
state $b$. Go to state $c$.

The third element is 2 because $2 \cdot U_1 = 2 \cdot 3 \le 7$, this leaves
$7 - (2 \cdot 3) = 1$ to be expanded. Go to state $c$ (follow label 2).

The last element is 1.
We conclude that this automaton can expand the number 41 correctly with the
given algorithm.
\end{example}

Observe that the generalized Zeckendorf expansion does not work in this case. 
The greedy algorithm would have expanded the number as 2100, but this is not 
accepted by the automaton.

\section{Survey}
Now we look at which substitutions are numeration-automatic. We have already 
seen that the Fibonacci automaton and the $k$-automata have this property.

\paragraph{The extended Fibonacci automata}
Figure \ref{fig:semi2} gives another example of a substitution of which we
shall prove that it is numeration-automatic.

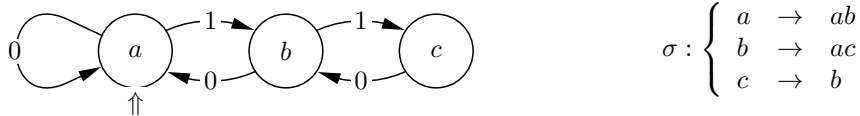
\begin{example}
\begin{figure}[H]
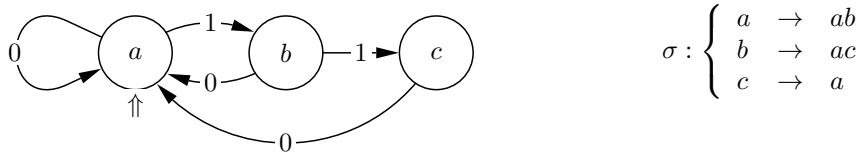

\vbox{
\begin{eqnarray*}
\spaces
\sigma: \left\{ \begin{array}{lll}
a &\rightarrow& ab\\
b &\rightarrow& ac\\
c &\rightarrow& b
\end{array} \right.
\end{eqnarray*}
\begin{graph}(0, 1)(-4, -2)
  \graphnodecolour{1}
  \graphnodesize{1}
  \roundnode{s1}(-2, 0) \nodetext{s1}(0, 0){$a$}
  \roundnode{s2}(0, 0)  \nodetext{s2}(0, 0){$b$}
  \roundnode{s3}(2, 0)  \nodetext{s3}(0, 0){$c$}

  \dirloopedge{s1}{50}(-1, 0) \freetext(-3.6, 0){0}
  \dirbow{s1}{s2}{.2} \bowtext{s1}{s2}{.2}{1}
  \dirbow{s2}{s1}{.2} \bowtext{s2}{s1}{.2}{0}
  \dirbow{s2}{s3}{.2} \bowtext{s2}{s3}{.2}{1}
  \dirbow{s3}{s2}{.2} \bowtext{s3}{s2}{.2}{0}

  \freetext(-2, -0.7){$\Uparrow$}
\end{graph} $(0 + (1 (10)^* 0))^* (\epsilon + (1 (10)^* (\epsilon + 1)))$}
\caption{A numeration-automaton}
\label{fig:semi2}
\end{figure}
\end{example}

\begin{figure}[H]
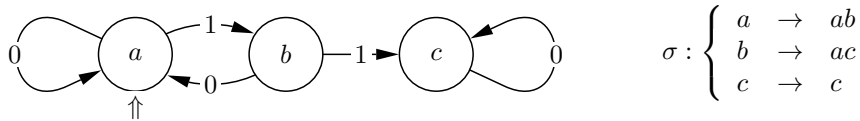

\vbox{
\begin{eqnarray*}
\spaces
\sigma: \left\{ \begin{array}{lll}
a &\rightarrow& ab\\
b &\rightarrow& ac\\
c &\rightarrow& a
\end{array} \right.
\end{eqnarray*}
\begin{graph}(0, 0)(-4, -1)
  \graphnodecolour{1}
  \graphnodesize{1}
  \roundnode{s1}(-2, 0) \nodetext{s1}(0, 0){$a$}
  \roundnode{s2}(0, 0)  \nodetext{s2}(0, 0){$b$}
  \roundnode{s3}(2, 0)  \nodetext{s3}(0, 0){$c$}

  \dirloopedge{s1}{50}(-1, 0) \freetext(-3.6, 0){0}
  \dirbow{s1}{s2}{.2} \bowtext{s1}{s2}{.2}{1}
  \dirbow{s2}{s1}{.2} \bowtext{s2}{s1}{.2}{0}
  \diredge{s2}{s3} \edgetext{s2}{s3}{1}
  \dirbow{s3}{s1}{.3} \bowtext{s3}{s1}{.3}{0}

  \freetext(-2, -0.7){$\Uparrow$}
\end{graph}}
\caption{A numeration-automaton}
\label{fig:semi3}
\end{figure}

\begin{figure}[H]
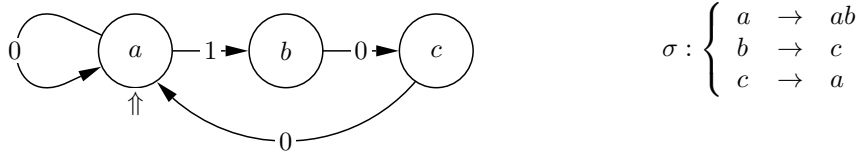

\vbox{
\begin{eqnarray*}
\spaces
\sigma: \left\{ \begin{array}{lll}
a &\rightarrow& ab\\
b &\rightarrow& c\\
c &\rightarrow& a
\end{array} \right.
\end{eqnarray*}
\begin{graph}(0, 0)(-4, -1)
  \graphnodecolour{1}
  \graphnodesize{1}
  \roundnode{s1}(-2, 0) \nodetext{s1}(0, 0){$a$}
  \roundnode{s2}(0, 0)  \nodetext{s2}(0, 0){$b$}
  \roundnode{s3}(2, 0)  \nodetext{s3}(0, 0){$c$}

  \dirloopedge{s1}{50}(-1, 0) \freetext(-3.6, 0){0}
  \diredge{s1}{s2} \edgetext{s1}{s2}{1}
  \diredge{s2}{s3} \edgetext{s2}{s3}{0}
  \dirbow{s3}{s1}{.3} \bowtext{s3}{s1}{.3}{0}

  \freetext(-2, -0.7){$\Uparrow$}
\end{graph}}
\caption{A numeration-automaton}
\label{fig:semi5}
\end{figure}

From Figure \ref{fig:semi2} on we will omit the regular expressions because it 
is tedious work and because the automaton gives a more insightful picture of 
the language than the regular expression.\\
\\
Just like the Fibonacci automaton, this one has a couple of `brothers and 
sisters', the most important of which are shown in Figures \ref{fig:semi3}, 
\ref{fig:semi4} and \ref{fig:semi5}.

\begin{figure}[H]
\vbox{
\begin{eqnarray*}
\spaces
\sigma: \left\{ \begin{array}{lll}
a &\rightarrow& ab\\
b &\rightarrow& ac\\
c &\rightarrow& c
\end{array} \right.
\end{eqnarray*}
\begin{graph}(0, 0)(-4, -1)
  \graphnodecolour{1}
  \graphnodesize{1}
  \roundnode{s1}(-2, 0) \nodetext{s1}(0, 0){$a$}
  \roundnode{s2}(0, 0)  \nodetext{s2}(0, 0){$b$}
  \roundnode{s3}(2, 0)  \nodetext{s3}(0, 0){$c$}

  \dirloopedge{s1}{50}(-1, 0) \freetext(-3.6, 0){0}
  \dirbow{s1}{s2}{.2} \bowtext{s1}{s2}{.2}{1}
  \dirbow{s2}{s1}{.2} \bowtext{s2}{s1}{.2}{0}
  \diredge{s2}{s3} \edgetext{s2}{s3}{1}
  \dirloopedge{s3}{50}(1,0) \freetext(3.6, 0){0}

  \freetext(-2, -0.7){$\Uparrow$}
\end{graph}}
\caption{A numeration-automaton}
\label{fig:semi4}
\end{figure}

\begin{figure}[H]
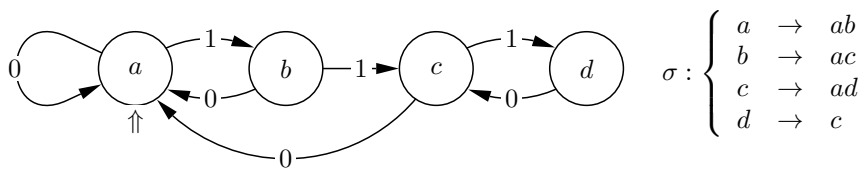

\vbox{
\begin{eqnarray*}
\spaces
\sigma: \left\{ \begin{array}{lll}
a &\rightarrow& ab\\
b &\rightarrow& ac\\
c &\rightarrow& ad\\
d &\rightarrow& c
\end{array} \right.
\end{eqnarray*}
\begin{graph}(0, 0)(-4, -1.3)
  \graphnodecolour{1}
  \graphnodesize{1}
  \roundnode{s1}(-2, 0) \nodetext{s1}(0, 0){$a$}
  \roundnode{s2}(0, 0)  \nodetext{s2}(0, 0){$b$}
  \roundnode{s3}(2, 0)  \nodetext{s3}(0, 0){$c$}
  \roundnode{s4}(4, 0)  \nodetext{s4}(0, 0){$d$}

  \dirloopedge{s1}{50}(-1, 0) \freetext(-3.6, 0){0}
  \dirbow{s1}{s2}{.2} \bowtext{s1}{s2}{.2}{1}
  \dirbow{s2}{s1}{.2} \bowtext{s2}{s1}{.2}{0}
  \diredge{s2}{s3} \edgetext{s2}{s3}{1}
  \dirbow{s3}{s1}{.3} \bowtext{s3}{s1}{.3}{0}
  \dirbow{s3}{s4}{.2} \bowtext{s3}{s4}{.2}{1}
  \dirbow{s4}{s3}{.2} \bowtext{s4}{s3}{.2}{0}

  \freetext(-2, -0.7){$\Uparrow$}
\end{graph}}
\caption{A numeration-automaton}
\label{fig:semi6}
\end{figure}

We can keep adding states as shown in Figure \ref{fig:semi6}

\begin{figure}[H]
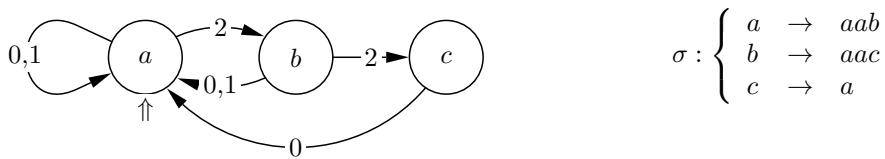

\vbox{
\begin{eqnarray*}
\spaces
\sigma: \left\{ \begin{array}{lll}
a &\rightarrow& aab\\
b &\rightarrow& aac\\
c &\rightarrow& a
\end{array} \right.
\end{eqnarray*}
\begin{graph}(0, 0)(-4, -1)
  \graphnodecolour{1}
  \graphnodesize{1}
  \roundnode{s1}(-2, 0) \nodetext{s1}(0, 0){$a$}
  \roundnode{s2}(0, 0)  \nodetext{s2}(0, 0){$b$}
  \roundnode{s3}(2, 0)  \nodetext{s3}(0, 0){$c$}

  \dirloopedge{s1}{50}(-1, 0) \freetext(-3.6, 0){0,1}
  \dirbow{s1}{s2}{.2} \bowtext{s1}{s2}{.2}{2}
  \dirbow{s2}{s1}{.2} \bowtext{s2}{s1}{.2}{0,1}
  \diredge{s2}{s3} \edgetext{s2}{s3}{2}
  \dirbow{s3}{s1}{.3} \bowtext{s3}{s1}{.3}{0}

  \freetext(-2, -0.7){$\Uparrow$}
\end{graph}}
\caption{A numeration-automaton}
\label{fig:semi7}
\end{figure}

We can further increase the number of transitions. See e.g. Figure 
\ref{fig:semi7}.

And here we stumble upon a class of automata which have incidence matrices of 
the form
\begin{displaymath}
\left( \begin{array}{cccccc}
x_{0, 0} & x_{0, 1} & x_{0, 2} & x_{0, 3} & \ldots & x_{0, n}\\
1 & 0 & 0 & 0 & \ldots & x_{1, n}\\
0 & 1 & 0 & 0 & \ldots & x_{2, n}\\
0 & 0 & 1 & 0 & \ldots & x_{3, n}\\
\vdots & \vdots & \vdots & \ddots & \ldots & \vdots\\
0 & 0 & 0 & \ldots & 1 & x_{n, n}
\end{array} \right)
\end{displaymath}
Here $x_{0, 0}$ must be larger or equal to 1 and
$x_{0, 1}, x_{0, 1} \ldots x_{0, n}$ and $x_{1, n}, x_{2, n} \ldots x_{n, n}$ 
may be any value between 0 and $k^\mathrm{max}$.

\begin{figure}[H]
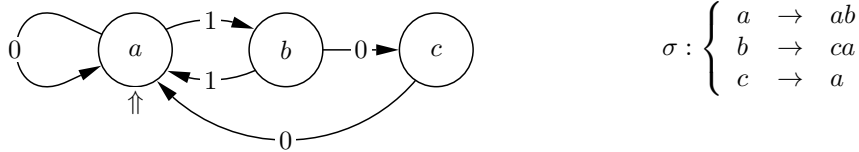

\vbox{
\begin{eqnarray*}
\spaces
\sigma: \left\{ \begin{array}{lll}
a &\rightarrow& ab\\
b &\rightarrow& ca\\
c &\rightarrow& a\\
\end{array} \right.
\end{eqnarray*}
\begin{graph}(0, 0)(-4, -1)
  \graphnodecolour{1}
  \graphnodesize{1}
  \roundnode{s1}(-2, 0) \nodetext{s1}(0, 0){$a$}
  \roundnode{s2}(0, 0)  \nodetext{s2}(0, 0){$b$}
  \roundnode{s3}(2, 0)  \nodetext{s3}(0, 0){$c$}

  \dirloopedge{s1}{50}(-1, 0) \freetext(-3.6, 0){0}
  \dirbow{s1}{s2}{.2} \bowtext{s1}{s2}{.2}{1}
  \dirbow{s2}{s1}{.2} \bowtext{s2}{s1}{.2}{1}
  \diredge{s2}{s3} \edgetext{s2}{s3}{0}
  \dirbow{s3}{s1}{.3} \bowtext{s3}{s1}{.3}{0}

  \freetext(-2, -0.7){$\Uparrow$}
\end{graph}}
\caption{A non-numeration-automaton}
\label{fig:nonsemi}
\end{figure}

Having an incidence matrix of this form is not sufficient, because the 
substitution in Figure \ref{fig:nonsemi} has such a matrix, but is not 
numeration-automatic.

The expansion sequence is as follows:
\begin{displaymath}
\{1, 2, 4, 7, 13, 24, 44, 81, 149, 274, 504, 927, \ldots \}
\end{displaymath}
The automaton crashes when we try to expand the number 5.

\begin{theorem}[$\sigma_0$-automatism] \label{thm:semiaut}
Let $\sigma_0, \ldots, \sigma_m$ be a substitution and let $0 < i, j \le m$. If
\begin{displaymath}
\sigma_0 \rightarrow {\sigma_0}^+ \sigma_*
\end{displaymath}
and
\begin{displaymath}
\sigma_i \rightarrow {\sigma_0}^* \sigma_*
\end{displaymath}
then the associated automaton is numeration-automatic.
\end{theorem}
\begin{proof}
The numeration system $(U_i)_{i \in \mathbb{N}}$ of $\sigma$ is given by
$U_i = |\sigma^i(\iota)|$. 
Consider a level $\ell$ of the infinite tree associated with $\sigma$, and 
number the nodes from 0 to $|\sigma^\ell(\iota)| = U_\ell$.
Consider a string $a_{\ell - 1} \ldots a_1 a_0$ leading to state $x$ with 
number $n$ at level $\ell$.

Call the states that the path will encounter: $x_\ell, \ldots, x_1, x$. 
Each node $x_i$ has $a_i$ siblings to the left and all these siblings are
labeled by $\iota$. In the $i$ remaining steps these $a_i$ $\iota$'s generate
$a_i |\sigma^i(\iota)|$ symbols to the left of $x$. The total number is 
\begin{displaymath}
\sum_{i = 0}^{\ell - 1} a_i |\sigma^i(\iota)| = \sum_{i = 0}^{\ell - 1} a_i U_i
\end{displaymath}
Thus the representation of $n$ by the automaton is $a_{\ell - 1} \ldots a_0$
too.
\end{proof}

\section{Combining automata}
The automata described above are `basic' automata, which means that the 
automata define their own numeration system and that the automata are minimal.
With minimal we mean that there is no automaton that defines the same 
numeration system, but has less states.
We can use these automata as a basis for other automata.\\
\\
\paragraph{The product automaton}
Let $A$ and $B$ be automata, let $\iota$ be the initial state of an automaton.
Let $\mathcal{S}_A$ be the states of $A$ and $\mathcal{S}_B$ the states of $B$.
We define a \emph{superstate} as a state consisting of a tuple $(a, b)$ with
$a \in \mathcal{S}_A$ and $b \in \mathcal{S}_B$. We make the product automaton
as follows.

\begin{itemize}
\item Start in the superstate $(\iota, \iota)$ and make this state the 
      initial state.
\item If both $\mathcal{A}$ and $\mathcal{B}$ have a transition from $\iota$ 
      labeled $i$, make a new superstate $(a, b)$ (the endpoints of the 
      transitions in both automata) and make a new transition from 
      $(\iota, \iota)$ to $(a, b)$ labeled $i$.
\item Do the same for all other states.
\end{itemize}

\begin{example}
\begin{figure}[H]
\begin{graph}(0, 3)(-4, -1.5)
  \graphnodecolour{1}
  \graphnodesize{1}
  \roundnode{s1}(-2, 0) \nodetext{s1}(0, 0){$a$}
  \roundnode{s2}(0, 0)  \nodetext{s2}(0, 0){$b$}

  \dirloopedge{s1}{50}(-1, 0) \freetext(-3.6, 0){0}
  \dirbow{s1}{s2}{.2} \bowtext{s1}{s2}{.2}{1}
  \dirbow{s2}{s1}{.2} \bowtext{s2}{s1}{.2}{0}

  \freetext(-2, -0.7){$\Uparrow$}
  \freetext(1, 0){$\times$}
\end{graph}
\begin{graph}(0, 3)(-9, -1.5)
  \graphnodecolour{1}
  \graphnodesize{1}
  \roundnode{s1}(-2, 0) \nodetext{s1}(0, 0){$a$}
  \roundnode{s2}(0, 0)  \nodetext{s2}(0, 0){$b$}

  \dirloopedge{s1}{50}(-1, 0) \freetext(-3.6, 0){0}
  \dirbow{s1}{s2}{.2} \bowtext{s1}{s2}{.2}{1}
  \dirbow{s2}{s1}{.2} \bowtext{s2}{s1}{.2}{1}
  \dirloopedge{s2}{50}(1, 0) \freetext(1.6, 0){0}

  \freetext(-2, -0.7){$\Uparrow$}
\end{graph}

\begin{graph}(0, 4)(-4, -3)
  \graphnodecolour{1}
  \graphnodesize{1}
  \textnode{s1}(-2, -2){$(a, a)$}
  \textnode{s2}(0, -2){$(b, b)$}
  \textnode{s3}(0, 0){$(a, b)$}
  \textnode{s4}(-2, 0){$(b, a)$}

  \dirloopedge{s1}{50}(-0.7, -0.7) \freetext(-3.1, -3.1){0}
  \diredge{s1}{s2} \edgetext{s1}{s2}{1}
  \diredge{s2}{s3} \edgetext{s2}{s3}{0}
  \dirloopedge{s3}{50}(0.7, 0.7) \freetext(1.1, 1.1){0}
  \diredge{s3}{s4} \edgetext{s3}{s4}{1}
  \diredge{s4}{s1} \edgetext{s4}{s1}{0}

  \freetext(-1.9, -2.45){$\Uparrow$}
  \freetext(-3, -1){=}
\end{graph}
\begin{graph}(0, 4)(-9, -3)
  \graphnodecolour{1}
  \graphnodesize{1}
  \roundnode{s1}(-2, -2) \nodetext{s1}(0, 0){$a$}
  \roundnode{s2}(0, -2)  \nodetext{s2}(0, 0){$b$}
  \roundnode{s3}(0, 0) \nodetext{s3}(0, 0){$c$}
  \roundnode{s4}(-2, 0)  \nodetext{s4}(0, 0){$d$}

  \dirloopedge{s1}{50}(-0.7, -0.7) \freetext(-3.1, -3.1){0}
  \diredge{s1}{s2} \edgetext{s1}{s2}{1}
  \diredge{s2}{s3} \edgetext{s2}{s3}{0}
  \dirloopedge{s3}{50}(0.7, 0.7) \freetext(1.1, 1.1){0}
  \diredge{s3}{s4} \edgetext{s3}{s4}{1}
  \diredge{s4}{s1} \edgetext{s4}{s1}{0}

  \freetext(-2, -2.7){$\Uparrow$}
  \freetext(-3, -1){=}
\end{graph}
\caption{Combined automata}
\label{fig:combined}
\end{figure}

In Figure \ref{fig:combined} we have combined the Fibonacci automaton with the 
Prouhet-Thue-Morse automaton by applying the product construction to the two 
automata.

The result is an automaton that generates the Prouhet-Thue-Morse sequence in 
the Fibonacci numeration system. By using the following exit map (projection
on the second coordinate):
\begin{eqnarray*}
\varphi: \left\{ \begin{array}{lll}
a &\rightarrow& a\\
b &\rightarrow& b\\
c &\rightarrow& b\\
d &\rightarrow& a
\end{array} \right.
\end{eqnarray*}
We obtain the Prouhet-Thue-Morse sequence again, and by using this exit map
(projection on the first coordinate):
\begin{eqnarray*}
\varphi: \left\{ \begin{array}{lll}
a &\rightarrow& a\\
b &\rightarrow& b\\
c &\rightarrow& a\\
d &\rightarrow& b
\end{array} \right.
\end{eqnarray*}
We get the Fibonacci sequence again.
\end{example}

Combining automata can result in a substitution for which it is not directly
clear that it is numeration-automatic. 

\begin{example} \label{ex:notclear}
When we combine the substitutions
\begin{eqnarray*}
\;\;\;
\sigma_A: \left\{ \begin{array}{lll}
a &\rightarrow& ab\\
b &\rightarrow& ac\\
c &\rightarrow& cc
\end{array} \right.
\end{eqnarray*}
with
\begin{eqnarray*}
\;\;\;
\sigma_B: \left\{ \begin{array}{lll}
a &\rightarrow& ab\\
b &\rightarrow& c\\
c &\rightarrow& ac
\end{array} \right.
\end{eqnarray*}
we obtain (after renaming $a := (a, a), b := (b, b), c := (a, c), d := (b, c), 
e := (c, c), f := (c, a), g := (c, b)$)
\begin{eqnarray*}
\sigma_{A \times B}: \left\{ \begin{array}{lll}
a &\rightarrow& ab\\
b &\rightarrow& c\\
c &\rightarrow& ad\\
d &\rightarrow& ae\\
e &\rightarrow& fe\\
f &\rightarrow& fg\\
g &\rightarrow& e
\end{array} \right.
\end{eqnarray*}
and the induced incidence matrix is not of the previously defined form
\begin{displaymath} \left( \begin{array}{ccccccc}
1 & 0 & 1 & 1 & 0 & 0 & 0 \\
1 & 0 & 0 & 0 & 0 & 0 & 0 \\
0 & 1 & 0 & 0 & 0 & 0 & 0 \\
0 & 0 & 1 & 0 & 0 & 0 & 0 \\
0 & 0 & 0 & 1 & 1 & 0 & 1 \\
0 & 0 & 0 & 0 & 1 & 1 & 0 \\
0 & 0 & 0 & 0 & 0 & 1 & 0 
\end{array} \right). \end{displaymath}
Still this substitution is numeration-automatic, because this automaton has 
exactly the same behavior as its `parents'.
\end{example}

\begin{theorem}[Product of $\sigma_0$-automata]
The product automaton of $\sigma_0$-automata is $\sigma_0$-automatic.
\end{theorem}
\begin{proof}
Every state $x$ has $|t(x)|$ outgoing transitions, the first $|t(x)| - 1$ of 
them are pointing to the initial state. We determine the product of two states 
$x$ and $y$. Assume that $|t(x)| \le |t(y)|$, this will result in a node with 
the first $|t(x)| - 1$ transitions pointing to $(\iota, \iota)$. Therefore
the resulting automaton is $\sigma_0$-automatic.
\end{proof}

\begin{theorem}[Product of $\sigma_0$- and $k$-automata]
The product automaton of a $\sigma_0$-automaton and a $k$-automaton is
numeration-automatic.
\end{theorem}
\begin{proof}
If the $k^{\mathrm{max}}$ of the $\sigma_0$-automaton is larger than the $k$ of
the $k$-automaton, then the $k^{\mathrm{max}}$ of the product automaton will
be $k$. Thus, when we leave out all transitions higher than $k$ in the
original $\sigma_0$-automaton, the resulting product automaton will be the 
same. Therefore we may assume that $k^{\mathrm{max}} \le k$. 

Consider the computation tree of the $\sigma_0$-automaton. When we apply the 
product construction to this tree, the structure of the tree does not change 
but the states are re-labeled. The first coordinate indicate the original
state. Thus Theorem \ref{thm:semiaut} still applies. The fixed point of the 
substitution is still computed correctly by the automaton because of Lemma 
\ref{lem:automata_substitutions}.
\end{proof}

The product automaton of two $k$-automata is $k$-automatic. If for example 
we construct the product automaton of a 2- and a 3-automaton, the result will
be a 2-automaton.

\section{Reverse reading}
We know from the theory of $k$-automata \cite{Fogg}, page 15 that if a 
$k$-automaton in direct reading exists, there also exists a $k$-automaton in 
reverse reading that accepts the same input language and gives the same 
mapping to the output alphabet. The only difference is that the automaton in
reverse reading reads the elements of its input from right to left instead of 
the normal order. The proof of this relies upon the existence of a $k$-kernel.

In general we can not make a $k$-kernel. However, we can construct an automaton
in reverse reading from an automaton in direct reading without having to 
construct a kernel.

We know this is possible because the theory of formal languages \cite{Wood}, 
page 419 states that a regular language is closed under the operation of 
mirroring, but this theory does not give an algorithm to make such an 
automaton. This is because the theory of formal languages does not apply to 
$k$-automata and numeration-automata directly. For example, the Prouhet-Thue-Morse 
automaton should be reduced to an automaton with one state and two loops with 
labels 0 and 1 according to this theory, because this is the minimal automaton 
that accepts $\{0, 1\}^*$. Moreover, all 2-automata should be reduced to this 
automaton.

\paragraph{Reversing an automaton}
To reverse an automaton, we only have to make all final states initial states
and vice versa and we must change the direction of the transitions. By doing 
this, we probably end up with a non-deterministic automaton. 
Fortunately, the non-deterministic automaton can be converted to a 
deterministic one using the \emph{subset construction} \cite{Wood}, page 118. 
This construction does not take into account that the set of final states may 
have partitions. Since our automata have output in their final states, the 
output induces a partition of the set of final states. General automata do 
not have an output in their states, a state is simply a final state or not.

Analogous to the subset construction we make our automaton in reverse reading, 
but we take into account the possibility of partitions by including the 
output function in the state.

\begin{example}
We have an automaton shown in Figure \ref{fig:det_direct}. Since the automata
we are going to make do not necessarily consist of final states only, we shall
mark the final states in the following automata with an inner circle.
\begin{itemize}
\begin{figure}[H]
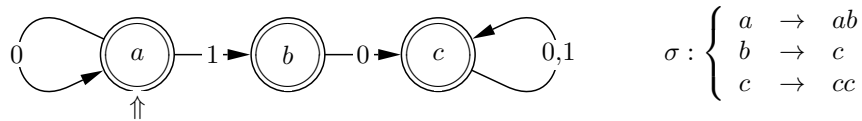

\vbox{
\begin{eqnarray*}
\spaces
\sigma: \left\{ \begin{array}{lll}
a &\rightarrow& ab\\
b &\rightarrow& c\\
c &\rightarrow& cc
\end{array} \right.
\end{eqnarray*}
\begin{graph}(0, 0)(-4, -1)
  \graphnodecolour{1}
  \graphnodesize{1}
  \roundnode{s1}(-2, 0) 
    \nodetext{s1}(0, -0.22){\circle{0.8}} \nodetext{s1}(0, 0){$a$}
  \roundnode{s2}(0, 0)  
    \nodetext{s2}(0, -0.22){\circle{0.8}} \nodetext{s2}(0, 0){$b$}
  \roundnode{s3}(2, 0)  
    \nodetext{s3}(0, -0.22){\circle{0.8}} \nodetext{s3}(0, 0){$c$}

  \dirloopedge{s1}{50}(-1, 0) \freetext(-3.6, 0){0}
  \diredge{s1}{s2} \edgetext{s1}{s2}{1}
  \diredge{s2}{s3} \edgetext{s2}{s3}{0}
  \dirloopedge{s3}{50}(1, 0) \freetext(3.6, 0){0,1}

  \freetext(-2, -0.7){$\Uparrow$}
\end{graph}}
\caption{A deterministic automaton in direct reading}
\label{fig:det_direct}
\end{figure}

\item First we swap final- and initial states and we change the direction of 
      the transitions as shown in Figure \ref{fig:nondet_reverse}.

\begin{figure}[H]
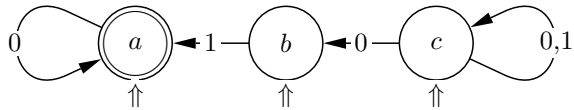

\begin{graph}(0, 2)(-4, -0.75)
  \graphnodecolour{1}
  \graphnodesize{1}
  \roundnode{s1}(-2, 0) 
    \nodetext{s1}(0, -0.22){\circle{0.8}} \nodetext{s1}(0, 0){$a$}
  \roundnode{s2}(0, 0) \nodetext{s2}(0, 0){$b$}
  \roundnode{s3}(2, 0) \nodetext{s3}(0, 0){$c$}

  \dirloopedge{s1}{50}(-1, 0) \freetext(-3.6, 0){0}
  \diredge{s2}{s1} \edgetext{s1}{s2}{1}
  \diredge{s3}{s2} \edgetext{s2}{s3}{0}
  \dirloopedge{s3}{50}(1, 0) \freetext(3.6, 0){0,1}

  \freetext(-2, -0.7){$\Uparrow$}
  \freetext(0, -0.7){$\Uparrow$}
  \freetext(2, -0.7){$\Uparrow$}
\end{graph}
\caption{A non-deterministic automaton in reverse reading}
\label{fig:nondet_reverse}
\end{figure}

\item Now we shall apply the subset construction \cite{Wood}, page 120
      on each of the initial states as if we were dealing with three automata. 
      We shall denote the state of the three automata in one state, so if
      the automata are in states $\{a\}, \{b\}, \{c\}$ respectively, we shall 
      notate this as $\{\{a\}, \{b\}, \{c\}\}$, because $a$ is the initial 
      state of the first automaton, $b$ the initial state of the second and 
      $c$ is the initial state of the third automaton.

\begin{figure}[H]
\begin{graph}(0, 2)(-4, -0.75)
  \graphnodecolour{1}
  \graphnodesize{1}
  \roundnode{s1}(-2, 0) 
    \nodetext{s1}(0, -0.22){\circle{0.8}} \nodetext{s1}(0, 0){$a$}
  \roundnode{s2}(0, 0) \nodetext{s2}(0, 0){$b$}
  \roundnode{s3}(2, 0) \nodetext{s3}(0, 0){$c$}

  \dirloopedge{s1}{50}(-1, 0) \freetext(-3.6, 0){0}
  \diredge{s2}{s1} \edgetext{s1}{s2}{1}
  \diredge{s3}{s2} \edgetext{s2}{s3}{0}
  \dirloopedge{s3}{50}(1, 0) \freetext(3.6, 0){0,1}

  \freetext(-2, -0.7){$\Uparrow$}
\end{graph}
\caption{Automaton 1}
\label{fig:nondet_reverse1}
\end{figure}
\begin{figure}[H]
\begin{graph}(0, 2)(-4, -0.75)
  \graphnodecolour{1}
  \graphnodesize{1}
  \roundnode{s1}(-2, 0) 
    \nodetext{s1}(0, -0.22){\circle{0.8}} \nodetext{s1}(0, 0){$a$}
  \roundnode{s2}(0, 0) \nodetext{s2}(0, 0){$b$}
  \roundnode{s3}(2, 0) \nodetext{s3}(0, 0){$c$}

  \dirloopedge{s1}{50}(-1, 0) \freetext(-3.6, 0){0}
  \diredge{s2}{s1} \edgetext{s1}{s2}{1}
  \diredge{s3}{s2} \edgetext{s2}{s3}{0}
  \dirloopedge{s3}{50}(1, 0) \freetext(3.6, 0){0,1}

  \freetext(0, -0.7){$\Uparrow$}
\end{graph}
\caption{Automaton 2}
\label{fig:nondet_reverse2}
\end{figure}
\begin{figure}[H]
\begin{graph}(0, 2)(-4, -0.75)
  \graphnodecolour{1}
  \graphnodesize{1}
  \roundnode{s1}(-2, 0) 
    \nodetext{s1}(0, -0.22){\circle{0.8}} \nodetext{s1}(0, 0){$a$}
  \roundnode{s2}(0, 0) \nodetext{s2}(0, 0){$b$}
  \roundnode{s3}(2, 0) \nodetext{s3}(0, 0){$c$}

  \dirloopedge{s1}{50}(-1, 0) \freetext(-3.6, 0){0}
  \diredge{s2}{s1} \edgetext{s1}{s2}{1}
  \diredge{s3}{s2} \edgetext{s2}{s3}{0}
  \dirloopedge{s3}{50}(1, 0) \freetext(3.6, 0){0,1}

  \freetext(2, -0.7){$\Uparrow$}
\end{graph}
\caption{Automaton 3}
\label{fig:nondet_reverse3}
\end{figure}

      Reading a 0 in state $\{\{a\}, \{b\}, \{c\}\}$ will result in state
      $\{\{a\}, \varnothing, \{b, c\}\}$, because automaton 1 stays in 
      state $a$ when reading a 0, automaton 2 crashes when reading a 0 in 
      state $b$ and automaton 3 goes to state $\{b, c\}$ because there are
      two outgoing branches labeled 0 in state $c$.

      We also have to determine the output function of each state. This is 
      done by observing in which coordinate the initial symbol (in this 
      case $a$) of the original automaton is.
      For example the state $\{\{a\}, \varnothing, \{b, c\}\}$ will have $a$
      as output and the state $\{\varnothing, \varnothing, \{a, b, c\}\}$ will
      have $c$ as output. This is because when we have reached the initial 
      state in reverse reading, we have also reached a final state in direct
      reading. For clarity, we write the output preceded by a slash. In the 
      first instance we write $\{\{a\}, \{b\}, \{c\} / a\}$ 
      If no output can be found, we use $\epsilon$ as output. This
      means that the state in question is not a final state.

      When we start in state $\{\{a\}, \{b\}, \{c\} / a\}$ and follow the
      labels 0 and 1, we get two new states. We now apply the same construction
      to these new states. The result is shown in Figure \ref{fig:det_reverse}.

\begin{figure}[H]
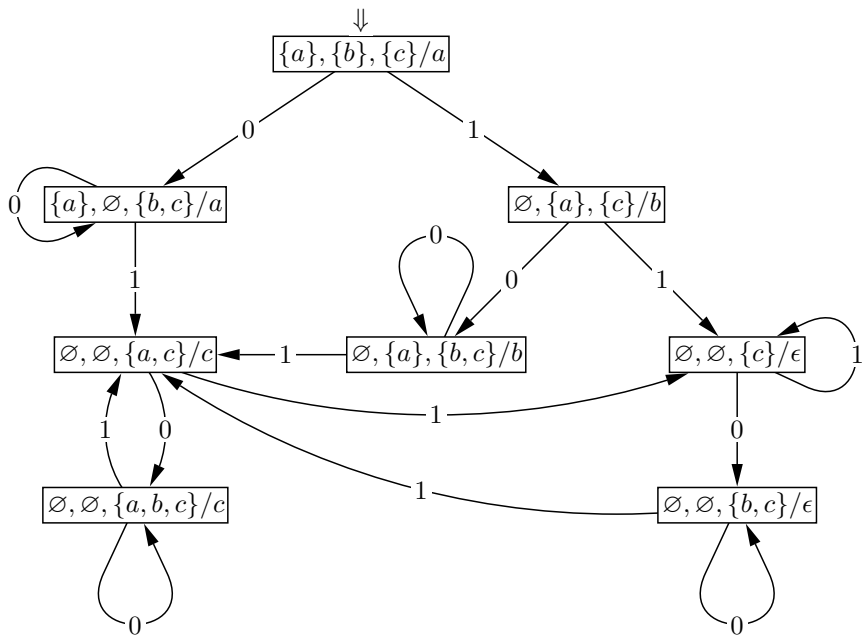

\begin{graph}(12, 8.5)(1, -0.5)
  \graphnodecolour{1}
  \textnode{s1}(6, 7){$\{a\}, \{b\}, \{c\} / a$}

  \textnode{s2}(3, 5){$\{a\}, \varnothing, \{b, c\} / a$}
  \textnode{s3}(9, 5){$\varnothing, \{a\}, \{c\} / b$}

  \textnode{s4}(3, 3){$\varnothing, \varnothing, \{a, c\} / c$}
  \textnode{s5}(7, 3){$\varnothing, \{a\}, \{b, c\} / b$}
  \textnode{s6}(11, 3){$\varnothing, \varnothing, \{c\} / \epsilon$}

  \textnode{s7}(3, 1){$\varnothing, \varnothing, \{a, b, c\} / c$}
  \textnode{s8}(11, 1){$\varnothing, \varnothing, \{b, c\} / \epsilon$}

  \diredge{s1}{s2} \edgetext{s1}{s2}{0}
  \diredge{s1}{s3} \edgetext{s1}{s3}{1}

  \dirloopedge{s2}{50}(-1, 0) \freetext(1.4, 5){0}
  \diredge{s2}{s4} \edgetext{s2}{s4}{1}
  \diredge{s3}{s5} \edgetext{s3}{s5}{0}
  \diredge{s3}{s6} \edgetext{s3}{s6}{1}

  \dirbow{s4}{s7}{.2} \bowtext{s4}{s7}{.2}{0}
  \dirbow{s4}{s6}{-.1} \bowtext{s4}{s6}{-.1}{1}
  \dirloopedge{s5}{50}(0, 1) \freetext(7, 4.6){0}
  \diredge{s5}{s4} \edgetext{s5}{s4}{1}
  \diredge{s6}{s8} \edgetext{s6}{s8}{0}
  \dirloopedge{s6}{50}(1, 0) \freetext(12.6, 3){1}
  
  \dirloopedge{s7}{50}(0, -1) \freetext(3, -0.6){0}
  \dirbow{s7}{s4}{.2} \bowtext{s7}{s4}{.2}{1}
  \dirloopedge{s8}{50}(0, -1) \freetext(11, -0.6){0}
  \dirbow{s8}{s4}{.1} \bowtext{s8}{s4}{.1}{1}

  \freetext(6, 7.45){$\Downarrow$}
\end{graph}
\caption{A deterministic automaton in reverse reading}
\label{fig:det_reverse}
\end{figure}

\item If we simplify the nodes to standard notation, we get the automaton 
      as shown in Figure \ref{fig:simpl_det_reverse}.

\begin{figure}[H]
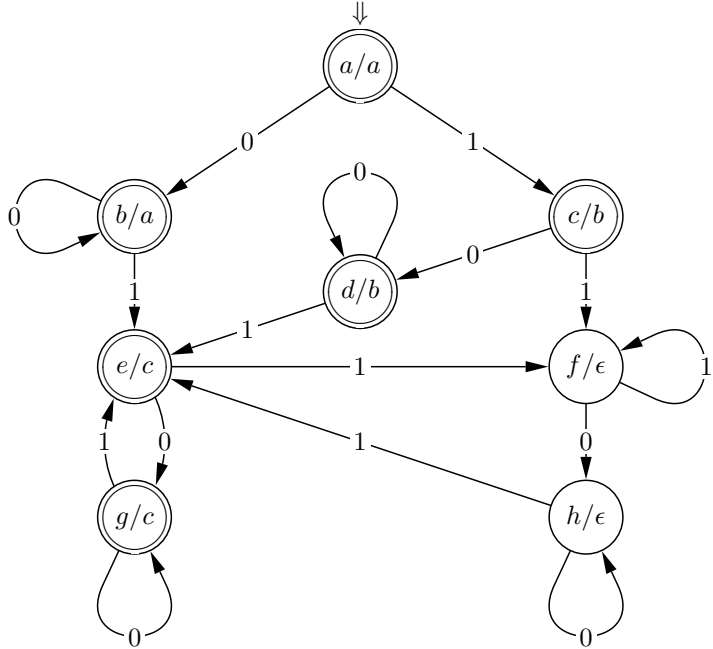

\begin{graph}(12, 8.5)(0, -0.5)
  \graphnodecolour{1}
  \graphnodesize{1}
  \roundnode{s1}(6, 7)
    \nodetext{s1}(0, -0.22){\circle{0.8}} \nodetext{s1}(0, 0){$a/a$}

  \roundnode{s2}(3, 5)
    \nodetext{s2}(0, -0.22){\circle{0.8}} \nodetext{s2}(0, 0){$b/a$}
  \roundnode{s3}(9, 5)
    \nodetext{s3}(0, -0.22){\circle{0.8}} \nodetext{s3}(0, 0){$c/b$}

  \roundnode{s5}(6, 4)
    \nodetext{s5}(0, -0.22){\circle{0.8}} \nodetext{s5}(0, 0){$d/b$}

  \roundnode{s4}(3, 3)
    \nodetext{s4}(0, -0.22){\circle{0.8}} \nodetext{s4}(0, 0){$e/c$}
  \roundnode{s6}(9, 3) \nodetext{s6}(0, 0){$f/\epsilon$}

  \roundnode{s7}(3, 1)
    \nodetext{s7}(0, -0.22){\circle{0.8}} \nodetext{s7}(0, 0){$g/c$}
  \roundnode{s8}(9, 1) \nodetext{s8}(0, 0){$h/\epsilon$}

  \diredge{s1}{s2} \edgetext{s1}{s2}{0}
  \diredge{s1}{s3} \edgetext{s1}{s3}{1}

  \dirloopedge{s2}{50}(-1, 0) \freetext(1.4, 5){0}
  \diredge{s2}{s4} \edgetext{s2}{s4}{1}
  \diredge{s3}{s5} \edgetext{s3}{s5}{0}
  \diredge{s3}{s6} \edgetext{s3}{s6}{1}

  \dirbow{s4}{s7}{.2} \bowtext{s4}{s7}{.2}{0}
  \diredge{s4}{s6} \edgetext{s4}{s6}{1}
  \dirloopedge{s5}{50}(0, 1) \freetext(6, 5.6){0}
  \diredge{s5}{s4} \edgetext{s5}{s4}{1}
  \diredge{s6}{s8} \edgetext{s6}{s8}{0}
  \dirloopedge{s6}{50}(1, 0) \freetext(10.6, 3){1}

  \dirloopedge{s7}{50}(0, -1) \freetext(3, -0.6){0}
  \dirbow{s7}{s4}{.2} \bowtext{s7}{s4}{.2}{1}
  \dirloopedge{s8}{50}(0, -1) \freetext(9, -0.6){0}
  \diredge{s8}{s4} \edgetext{s8}{s4}{1}

  \freetext(6, 7.7){$\Downarrow$}
\end{graph}
\caption{A simplified deterministic automaton in reverse reading}
\label{fig:simpl_det_reverse}
\end{figure}
\end{itemize}
\end{example}

\section{Numeration systems}
The class of numeration-automatic substitutions has an interesting subclass:
the subclass of substitutions that define a full numeration system. We 
believe that this is the class of $\sigma_0$-automata with the restriction
that the cardinality of the images of the substitution do not increase,
so $|\sigma_0| \ge |\sigma_1| \ge \ldots$. In this section we show that the
condition suffices.

Without loss of generality, we can write a $\sigma_0$-substitution in the
following way
\begin{eqnarray*}
\sigma_0       &\rightarrow& {\sigma_0}^+ \sigma_1\\
               &\ldots&\\
\sigma_{k - 1} &\rightarrow& {\sigma_0}^* \sigma_k\\
\sigma_k       &\rightarrow& {\sigma_0}^* \sigma_*
\end{eqnarray*}
We assume this special form throughout this section.

First we derive a lemma for general $\sigma_0$-automata.

\begin{lemma}[The recurrent function of a substitution.]
If $\sigma$ is a substitution of $\sigma_0$-automatic type with $k$ 
substitution rules, then the numeration system is a linear recurrent function 
of at most order $k$.
\end{lemma}
\begin{proof}
Since each substitution rule is of the form 
\begin{displaymath}
\sigma_i \rightarrow \sigma_0^{j_i} \sigma_{i + 1}, j_i \ge 0
\end{displaymath}
and the first rule is of the form
\begin{displaymath}
\sigma_0 \rightarrow \sigma_0^{j_0} \sigma_1, j_0 > 0,
\end{displaymath}

we can extract part of the recurrent function from the first substitution rule.
This results in a relation depending on $\sigma_0$ and $\sigma_1$: 
$a_n = j_0 a_{n - 1} + b_{n - 1}$. Hence $b_{n - 1}$ can be expressed as  
$a_n - j_0 a_{n - 1}$, analogously we
can write the next equation as $b_n = j_1 a_{n - 1} + c_{n - 1}$. This yields
$c_{n - 1} = b_n - j_1 a_{n - 1} = a_{n + 1} - j_0 a_n - j_1 a_{n - 1}$. In 
this way we can successively write $b_{n - 1}, c_{n - 1}, d_{n - 1}, \ldots$
as linear combinations of $a_n, a_{n + 1}, a_{n + 2}, \ldots$. Finally we get
a linear homogeneous recurrence relation with constant coefficients of the
numbers $a_n$. The order of this recurrence equals the number of substitution
rules.
\end{proof}

\begin{example}
Consider the following substitution scheme.
\begin{eqnarray*}
\sigma: \left\{ \begin{array}{lll}
a &\rightarrow& ab\\
b &\rightarrow& aac\\
c &\rightarrow& d\\
d &\rightarrow& ac
\end{array} \right.
\end{eqnarray*}
First we write the substitution rules as recurrent functions.
\begin{eqnarray*}
a_n &=& a_{n - 1} + b_{n - 1}\\
b_n &=& 2a_{n - 1} + c_{n - 1}\\
c_n &=& d_{n - 1}\\
d_n &=& a_{n - 1} + c_{n - 1}
\end{eqnarray*}

Now we start eliminating 
\begin{eqnarray*}
b_{n - 1} &=& a_n - a_{n - 1}\\
c_{n - 1} &=& b_n - 2a_{n - 1} = a_{n + 1} - a_n - 2a_{n - 1}\\
d_{n - 1} &=& c_n = a_{n + 2} - a_{n + 1} - 2a_n
\end{eqnarray*}
So
\begin{eqnarray*}
a_{n + 3} - a_{n + 2} - 2a_{n + 1} &=& 
  a_{n - 1} + a_{n + 1} - a_n - 2a_{n - 1}\\
a_{n + 3} &=& a_{n + 2} + 3a_{n + 1} - a_n - a_{n - 1}
\end{eqnarray*}
And the final result is:
$a_n = a_{n - 1} + 3a_{n - 2} - a_{n - 3} - a_{n - 4}$, which is the recurrence
that generates the sequence $1, 2, 5, 10, 22, 45, 96, 199, 420, 876, \ldots$
(with the appropriate initial conditions
$|a|, |\sigma(a)|, |\sigma^2(a)|, |\sigma^3(a)|$).
\end{example}

\begin{theorem}[Full numeration systems and numeration-automatism] 
\label{thm:fsa}
If $\sigma$ is a $\sigma_0$-substitution, with 
$|\sigma(\sigma_0)| \ge |\sigma(\sigma_1)| \ge \ldots \ge |\sigma(\sigma_k)|$, 
then the associated automaton generates a full numeration system.
\end{theorem}
\begin{proof}
Consider the computation tree associated with the substitution $\sigma$. Choose
$n$ such that $|\sigma^n(\sigma_0)| \le x < |\sigma^{n + 1}(\sigma_0)|$.

Let $a$ be a state with $\sigma(a) = \sigma_0^{\ell_a} a'$, with 
$a' \in \{\epsilon, \sigma_1, \ldots, \sigma_k\}$.

The state $a$ results in $|\sigma^{m + 1}(a)|$ states $m$ levels deeper in 
the tree. Because $|\sigma^m(\sigma_0)| = U_m$, these states can be partitioned
in $\ell_a$ sets of $U_m$ states and one set of $|\sigma(a')|$ states with
$|\sigma^m(a')| < U_m$ because $\sigma(a') < \sigma(\sigma_0)$.

We now iterate the following procedure starting with the triple 
$(x, \sigma_0, n)$.

Consider the triple $(x, a, m)$ with $x$ an integer such that $x < U_{m + 1}$
and $a$ a state. We write $x = \ell_m U_m + x'$ with $0 \le x'< U_m$. Then
$\ell_m < \frac{U_{m + 1}}{U_m} \le k^{\mathrm{max}} + 1$, hence 
$\ell_m \le k^{\mathrm{max}}$. If $\ell_m < \ell_a$, then $x$ is the result
of state $\sigma_0$ when we go $m$ levels higher. If $\ell_m = \ell_a$, then
$x$ is the result of state $a'$ when we go $m$ levels higher. In the former
case, we replace the triple $(x, a, m)$ by $(x', \sigma_0, m - 1)$, in the
latter case by $(x', a', m - 1)$. Recall that if $a' = \sigma_i, a = \sigma_j$,
then $i \ge j$ and therefore $|\sigma(\sigma_i)| \le |\sigma(\sigma_j)|$.

Obviously $x = \ell_n U_n + \ell_{n - 1} U_{n - 1} + \ldots + \ell_0 U_0$.
However, by construction we have that $\ell_n \ell_{n - 1} \ldots \ell_0$ is
the expansion of $x$ in the computation tree. Hence the numeration system is
full.
\end{proof}

We remark that the extra condition can not be dropped. See Example 
\ref{ex:non-semi}.

\section{Conclusion}
We have seen that with the automatic expansion we can successfully find 
numeration systems for a class of automata. The most important findings are
that the substitution defines a numeration system and that the automaton 
defines an expansion algorithm. Combine them and we get the class of 
numeration-automatic sequences.

The sequences which are not numeration-automatic remain interesting, because 
the associated automaton calculates most of the letters in the fixed point 
correctly, but it leaves gaps.  Maybe it is somehow possible to `repair' the 
automaton to correct this behavior, for example by using a stack automaton, 
but this is beyond the scope of this document.

\section{More on numeration-automatism}
The following website has a program that checks a substitution for 
numeration-automatism.
\begin{verbatim}
http://www.liacs.nl/~jlaros/semi/
\end{verbatim}

The ``On-Line Encyclopedia of Integer Sequences'' is a huge database of integer
sequences. The author has contributed some sequences and commented on some
other sequences. See
\begin{verbatim}
http://www.research.att.com/~njas/sequences/Seis.html
\end{verbatim}

for the main page, or one of the following pages.
\begin{verbatim}
http://www.research.att.com/cgi-bin/access.cgi/as/njas/sequences/
eisA.cgi?Anum=A000045
http://www.research.att.com/cgi-bin/access.cgi/as/njas/sequences/
eisA.cgi?Anum=A101197
http://www.research.att.com/cgi-bin/access.cgi/as/njas/sequences/
eisA.cgi?Anum=A101168
http://www.research.att.com/cgi-bin/access.cgi/as/njas/sequences/
eisA.cgi?Anum=A101169
http://www.research.att.com/cgi-bin/access.cgi/as/njas/sequences/
eisA.cgi?Anum=A101399
http://www.research.att.com/cgi-bin/access.cgi/as/njas/sequences/
eisA.cgi?Anum=A101400
\end{verbatim}

\section{Acknowledgements}
The author wishes to thank Dr. Hendrik Jan Hoogeboom and Prof. Dr. R. Tijdeman
for their numerous comments and insightful ideas.


\begin{thebibliography}{XX}
\bibitem{Fogg} Fogg, N. Pytheas. Substitutions is Dynamics, Arithmetics and
               Combinatorics, Springer Verlag, 2002.
\bibitem{Wood} Wood, Derick. Theory of Computation, John Wiley \& sons, inc, 
               1987.
\bibitem{Loth} Lothaire, M. Algebraic combinatorics on words, Cambridge, 2002.
\bibitem{Hol} Hollander, M. Greedy numeration systems and regularity.
              Theory Comput. Systems 31 (1998), 111--133.
\end{thebibliography}
\end{document}